\documentclass[letterpaper, 10 pt, conference]{ieeeconf}  

\IEEEoverridecommandlockouts                              

\usepackage{graphics} 
\usepackage{epsfig} 
\usepackage{mathptmx} 
\usepackage{times} 
\usepackage{amsmath} 
\usepackage{amssymb}  
\usepackage{bm}
\usepackage{algorithm}
\usepackage{multirow}
\usepackage{verbatim}
\usepackage{graphicx}
\usepackage{hyperref}
\usepackage{graphicx,fancyhdr}

\usepackage{booktabs} 

\usepackage{subfigure}
\usepackage{color}

\title{Autonomous Person-Specific Following Robot}

\author{Wesley P. Chan$^{1}$, Sina Radmard$^{2}$, Zhao Quan Hew$^{1}$, Jon Morris$^{2}$,
\\ Elizabeth Croft $^{1}$, and H.F. Machiel Van der Loos$^{3}$
\thanks{$^{1}$Monash University, $^{2}$JDQ Systems, $^{3}$University of British Columbia}

} 
\begin{document}

\maketitle
\thispagestyle{empty}
\pagestyle{empty}

\begin{abstract}
Following a specific user is a desired or even required capability for service robots in many human-robot collaborative applications. However, most existing person-following robots follow people without knowledge of who it is following. In this paper, we proposed an identity-specific person tracker, capable of tracking and identifying nearby people, to enable person-specific following. Our proposed method uses a Sequential Nearest Neighbour with Thresholding Selection algorithm we devised to fuse together an anonymous person tracker and a face recogniser. Experiment results comparing our proposed method with alternative approaches showed that our method achieves better performance in tracking and identifying people, as well as improved robot performance in following a target individual.

\end{abstract}

\section{INTRODUCTION}
        \label{sec:introduction}
        Recently, the area of service robotics has seen many developments and advances, with target applications ranging from shopping assistants \cite{iwamura2011} and last mile delivery \cite{Du2019}, to agricultural harvesting \cite{masuzawa2017development} and disaster response \cite{Burke2004}. In many of these applications, the ability to follow the user is a desirable, if not required, skill for the robot to have. In some scenarios, the ability to follow the user serves as a more intuitive and convenient method for repositioning or relocating the robot. For example, a search and rescue team leader leading the robot from one search area to the next search area. In other scenarios, the ability to follow the user composes the core functionality. For example, a shopping assistant following its user around the supermarket while carrying the shopping basket. 

Many person-following robots are described in the literature. A taxonomy of different types of person-following robots, including autonomous ground, air, and underwater robots, is given by \cite{islam2019person}, with a wide range of human-robot collaborative applications. While many applications require the robot to follow a specific person (the user), in many existing works on person-following robots, the user or experimenter would first initialise a tracking person/object/region \cite{Chen2017,Chen2007,Chivilo2004}, or the target person would be asked to wear a specific colour \cite{Siebert2020}, and the robot simply tracks and follows the initialised entity, or blob of colour, without knowing who or what it is following. This is because most existing methods use a person or object tracker for person-following, and most existing trackers do not distinguish the identity of each person \cite{linder2014,Ess2008,Scholer2011,Haselich2014}. This means that the robot is not able to identify which person is the user, and if the tracker drifts or loses track of the user, it cannot re-identify the user. To enable robots to more reliably and robustly follow their users in different applications, we propose an identity-specific person tracker, capable of tracking near by people and their identities. We implement our proposed method on a mobile robot, and compared its performance with alternative methods in an experiment involving a person-specific following task.
        
\section{RELATED WORKS}
        \label{sec:related_works}
        We provide a review of methods in the literature for two main topics related to enabling person-specific following - 1) Person tracking, for determining the spacial position of the surrounding people. 2) Person (re-)identification, for determining the identity of the surrounding people, or re-identify a lost following target. 

\subsection{Person Tracking}
While many existing algorithms are capable of tracking the position of multiple people and can run in realtime, they are not able to distinguish the identity of the tracked persons. Hence, person-following robot applications developed using these anonymous person trackers are not capable of knowing who they are following, or confirming if they are following the correct user. Inevitably, there will be instances of drift or target lost, and these person trackers typically cannot recover from such events. This results in the robot following the wrong person without knowing, or the robot not able to find the target person once it moves temporarily out of sight, or behind occlusion. Some algorithms rely on 2D laser sensors, using a clustering algorithm to first group together nearby point clusters, then using a trained classifier to detect human legs \cite{arras2007using,lu2013towards}. A nearest-neighbour algorithm is then used to track people from frame to frame. Approaches using recurrent neural networks to develop end-to-end systems for tracking people using 2D laser sensors have also been proposed \cite{ondruska2016end}. Similar methods utilising 3D laser sensors for detecting and tracking people using a classifier have also been developed \cite{yan2020online}. Aside from laser sensor, camera sensors are also used for people tracking utilising particle filters \cite{breitenstein2009robust}, and convoluted neural networks \cite{ma2018customized}. Algorithms using a combination of different sensors have also been developed. For example, \cite{linder2014} combined depth image and 2D laser data with template matching and classification algorithms for people tracking.

\subsection{Person (Re-)identification}
Current research in the identification of humans takes a variety of approaches and has been extensively studied \cite{mazzon2012person,bedagkar2014survey}. These approaches crudely boil down to the identification of two types of extracted features \cite{eisenbach2012view}. The first type being appearance-based features which includes attributes such as colour \cite{farenzena2010person,eisenbach2015user} or histogram of gradients (HOG) in an image frame \cite{dalal2005histograms}. The latter type of features are biometrics such as the face, gait \cite{koide2016identification}, body geometric appearance \cite{Kirchner2012}, and skeletal information \cite{munaro2014feature}. The benefit of using appearance-based approaches is the reduction of the feature space, therefore they can be much more real-time capable. However, relying on features such as colour can suffer from poor illumination conditions and observation noise. Identifying biometrics such as gait or skeletal information requires a full view of the body, which is impractical for human-robot interaction in proximity.

Still, biometrics are more consistently reliable and robust in the long term compared to appearance-based features. For example, a person may change their clothes, and a robot module that relies on colour features would not be able to identify that person anymore. Hence, we propose to use faces, as an easily observable biometric using typical robot sensors, for identifying individuals. Facial recognition has been researched for a long time \cite{turk1991face} and has been reliably implemented \cite{schroff2015facenet}. Plus, face recognition is arguably easy-to-implement and versatile as only images of the face is needed for registration and identification. Facial recognition algorithms robust to day-to-day changes such as glasses, hats, hair cuts are also available \cite{singh17,sabharwal19}.
       
\section{IDENTITY-SPECIFIC PERSON TRACKING}
        \label{sec:method}
        
\begin{figure*}[h]
\centering
\includegraphics[width=\textwidth]{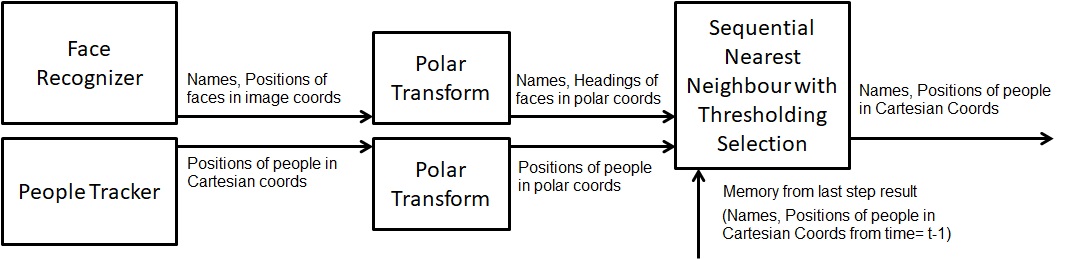}
\caption{Identity-specific people tracker pipeline.}
\label{fig:pipeline}
\end{figure*}

To enable identity-specific person tracking and following by robots, we propose a Sequential Nearest Neighbour with Thresholding Selection (SNNTS) algorithm (presented in the next section). The proposed algorithm utilises anonymised person-tracking and face recognition, fusing results from the two to achieve identity-specific person tracking. Figure \ref{fig:pipeline} gives an overview of our tracker pipeline. The following subsections explains each components in more detail.

\subsection{Person Tracker}
The person tracker we use combines the use of depth image and 2D laser scan data for tracking people. The 2D laser data is first passed into a clustering algorithm to cluster nearby groups of data points. The identified clusters are then passed into a trained classifier to detect persons by detecting legs. The depth image data is used to detect upper bodies of people using template matching. People detection results from 2D laser data and depth image data are aggregated. Then, people are tracked by using a nearest neighbour algorithm from frame to frame. The inter-frame velocity of each tracked person is estimated, and this estimated velocity is used to predict the positions of occluded persons up to a specified number of frames, using a constant velocity model. We used the person tracker implementation provided by \cite{linder2014}. The person tracker outputs a list of tracked (anonymous) persons as:
\begin{equation}
[x^p_i(t), y^p_i(t), z^p_i(t)].
\label{eq:person_tracker_output}
\end{equation}

\subsection{Person Identification}
\begin{figure}[t]
\centering
\includegraphics[width=.30\textwidth]{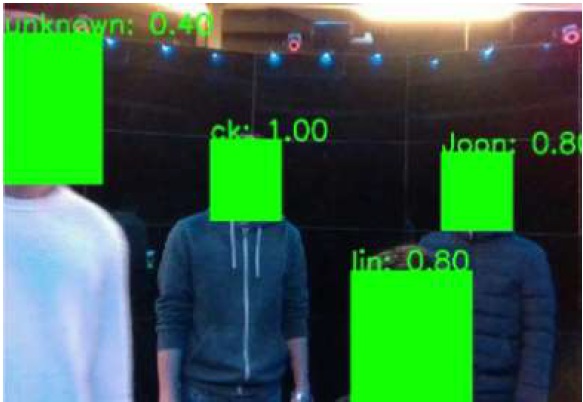}
\caption{Example face recognition results showing bounding boxes of detected faces, predicted identities (names), and face recognition scores. The name is shown as "unknown" when the face recognition score is lower than a set threshold.}
\label{fig:face_recognition_example}
\end{figure}

We use face recognition to identify nearby people. 
Our face recognition module encapsulates the face recognition
system Facenet \cite{schroff2015facenet}, and is derived from \cite{facenet}, an open-source
implementation of FaceNet. FaceNet itself is a system that can
verify and recognize faces through matching compact
embeddings generated using a deep convolutional neural
network. Our implementation of the method uses a pre-trained
model of the neural network, provided by \cite{facenet}. The architecture
of the model is the Inception ResNet v1 \cite{szegedy2016inception}, and the model was pre-trained with the dataset in \cite{yi2014learning}. Faces of the user(s) are given to train the face recogniser ahead of time, extracting the embeddings. During run-time, the input image is first passed to a multi-task convolutional neural network to detect regions of faces \cite{zhang2016joint}. These detected regions containing faces are then passed to the face recogniser to determine the identity through classifying the embedding of each detected face. Figure \ref{fig:face_recognition_example}
shows an example output of our face recogniser. The face recognition module outputs a list of persons recognised, with the corresponding positions of each recognised face: 
\begin{equation}
[name^f_j(t), u^f_j(t), v^f_j(t)],
\label{eq:face_recognition_output}
\end{equation}
where $u^f_j(t)$ and $v^f_j(t)$ are the pixel row and column of the corresponding bounding box center.

\subsection{Sequential Nearest Neighbour with Thresholding Selection (SNNTS)}
The person tracker outputs (anonymous) people positions in the world space (Eq. \ref{eq:person_tracker_output}), while the face recogniser outputs recognized people names and their location in image space (Eq. \ref{eq:face_recognition_output}). To be able to compare and integrate these results, we first convert each of them into polar coordinates to obtain the $\theta$ (heading) component, relative to the robot:
\begin{equation}
\theta^p_i(t)=atan2(y^p_i(t), x^p_i(t)),
\label{eq:person_tracker_polar}
\end{equation}
and
\begin{equation}
{\theta}^f_j(t)=atan2(u^f_j(t), f_x),
\label{eq:face_recognition_polar}
\end{equation}
where $f_x$ is the focal length in the horizontal axis from the camera calibration matrix.

Using the headings of the tracked persons and recognised faces given by Eq. \ref{eq:person_tracker_polar} and Eq. \ref{eq:face_recognition_polar} respectively, our proposed SNNTS algorithm first attempts to find a corresponding recognised face for each tracked person, by selecting the recognised face that has the closet heading as the tracked person. To reject improbable matches, the algorithm only considers potential correspondences that are within a threshold distance $\theta_{thres}$. At run-time. our algorithm also maintains a memory of tracked persons identified in the previous time step $t-1$. If a corresponding face is not found for a tracked person at the current time step $t$, then the previously identified name at $t-1$ is assumed. This enables our algorithm to continuously track and identify people even when the face recogniser failed to recognise their face temporarily (e.g.,
people turning their faces away from the robot). Our proposed SNNTS algorithm is given togehter in Algorithm 1 and Algorithm 2. Although $\theta_{thres}$ can be chosen to be distance-dependent, we set $\theta_{thres}$ to 15 degrees empirically in our experiment, as this was found to yield good performance. Our algorithm outputs a list of identified persons' name and position:
\begin{equation}
[name_n(t), x_n(t), y_n(t), z_n(t)],
\end{equation}
allowing the robot to track the position of each nearby person, and knowing who each person is. 

Our algorithm provides a much more robust and reliable method for identifying and tracking people compared to if we were to simply label the tracked persons of an anonymous tracker (which, we will compare with in our experiment). In real scenarios, person trackers often drift and there will be instances when the target is temporarily lost or occluded. A simple labeling approach would not be able to recover or re-identify individuals after such events. Furthermore, even if we were to simply attempt to label the anonymously tracked person with the face recognition results in each frame, often there are periods of time when a face recogniser is not able to recognise the individuals. This can be because of poor/variable lighting, (partial) occlusions, individuals moving to or standing at locations outside of the camera's field of view, or people not facing the robot. Our SNNTS algorithm overcomes these issues.

\begin{figure}[t]
\centering
\includegraphics[width=.48\textwidth]{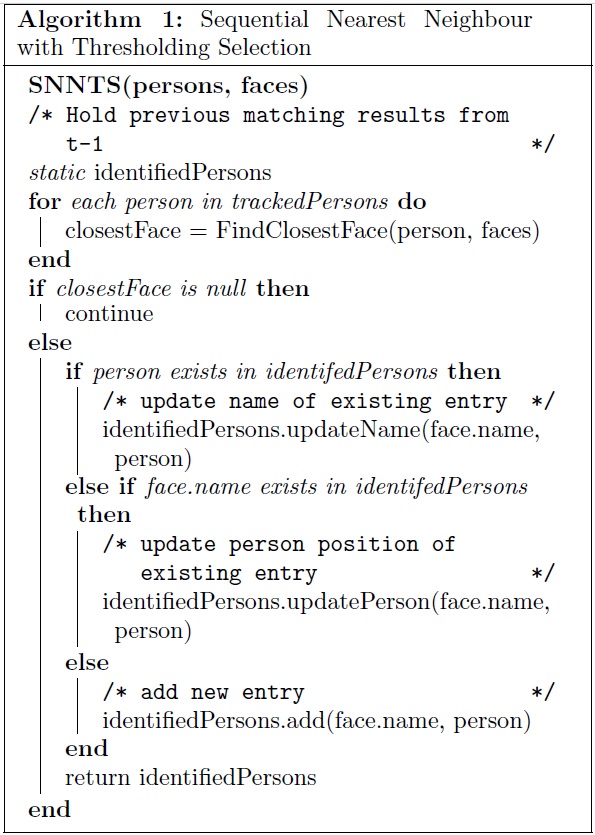}
\end{figure}

\begin{figure}[t]
\centering
\includegraphics[width=.48\textwidth]{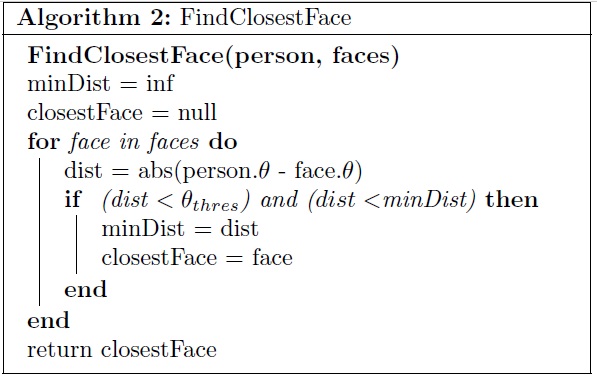}
\end{figure} 
        
\section{ROBOT HARDWARE}
        \label{sec:robot}
        \begin{figure}[t]
\centering
\includegraphics[width=.40\textwidth]{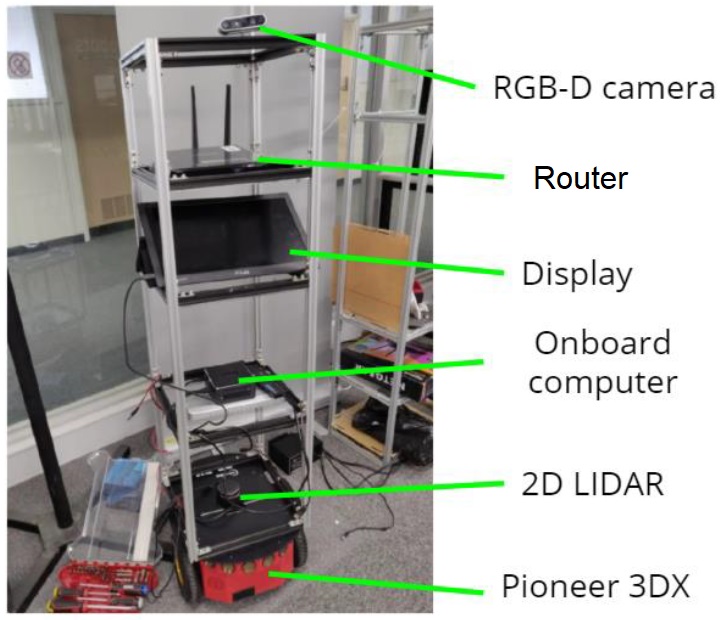}
\caption{Our robot hardware.}
\label{fig:robot}
\end{figure}

We implemented our algorithm on our mobile robot, shown in Figure \ref{fig:robot}, and tested our implementation with a set of person-following tasks.
Our robot platform comprises a Pioneer P3DX mobile robot, equipped with a 360 deg 2-D laser scanner (RPLIDAR A3) and an RGB-D camera (Intel RealSense Depth Camera D435). The computer running the face tracker has an NVIDIA GeForce GTX 1070 Max-Q GPU for its neural network operations.  
        
\section{EXPERIMENT}
        \label{sec:experiment}
\begin{figure}[b]
\centering
\includegraphics[width=.40\textwidth]{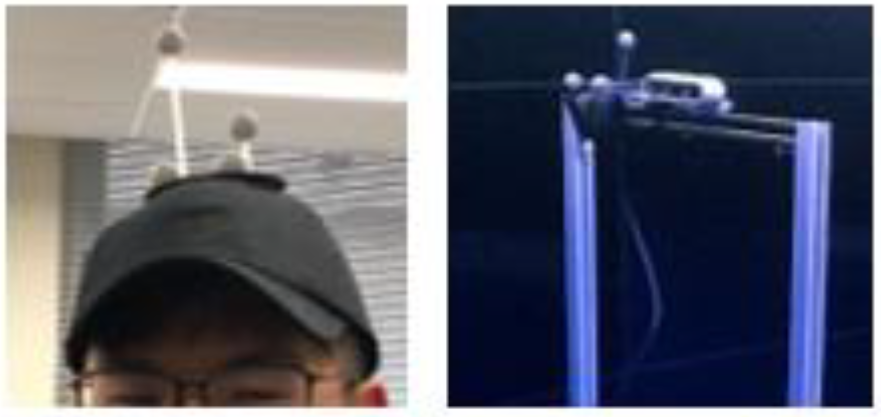}
\caption{Infrared reflective markers on robot and people for motion capture.}
\label{fig:marker}
\end{figure}

        We conducted several person-tracking and person-following experiments using a motion capture system (Vicon) to measure the ground truth positions of people and robot. Infrared reflective markers for motion capture are placed on the robot and the people as shown in Figure \ref{fig:marker}. 

We tested five experiment scenarios:
\begin{itemize}
    \item Exp 1: The robot and three people in a stationary position, facing the robot. (As a baseline.)
    \item Exp 2: The robot in a stationary position, with three people walking around the robot.
    \item Exp 3: The robot is set to look at/face a specified target person by controlling only its yaw, with three people walking around the robot.
    \item Exp 4: The robot is set to follow the closest tracked person, with three people walking around the robot.
    \item Exp 5: The robot is set to follow a specified target person, with three people walking around the robot.
\end{itemize}

We compared our proposed algorithm with two alternative methods. The first alternative method uses a face recogniser (FaceNet \cite{facenet}) in conjunction with a calibrated depth image. The second alternative method uses an anonymous person tracker (SPENCER \cite{linder2014}). These two alternative methods are elaborated below.

\textbf{FaceNet}. Our robot is equipped with an RGBD camera. The camera outputs an RGB image and a calibrated point cloud. FaceNet is used to recognise the faces in the RGB image. The corresponding region of each recognised faces in the point cloud data is then referenced. We clustered data points in the corresponding region of the point cloud, and the distance from the person (and hence, Cartesian location as well) can be estimated using the centroid of the cluster. This method has limited accuracy since it was observed that the point cloud data at object edges is often noisy, and may be fused with background objects.

\textbf{SPENCER}. The SPENCER tracker \cite{linder2014} provides the positions of the nearby people (but not their identity). For our experiment, we manually provided the identity of each person at the first frame. This method, however, means that if the tracker drifts or temporarily loses track of a person, it will never be able to re-identify the person or recover. In our experiment, we indeed observed drift and temporary target loss to occur frequently, due to the challenging experiment scenarios we presented.

\begin{figure}[t]
\centering
\includegraphics[width=.40\textwidth]{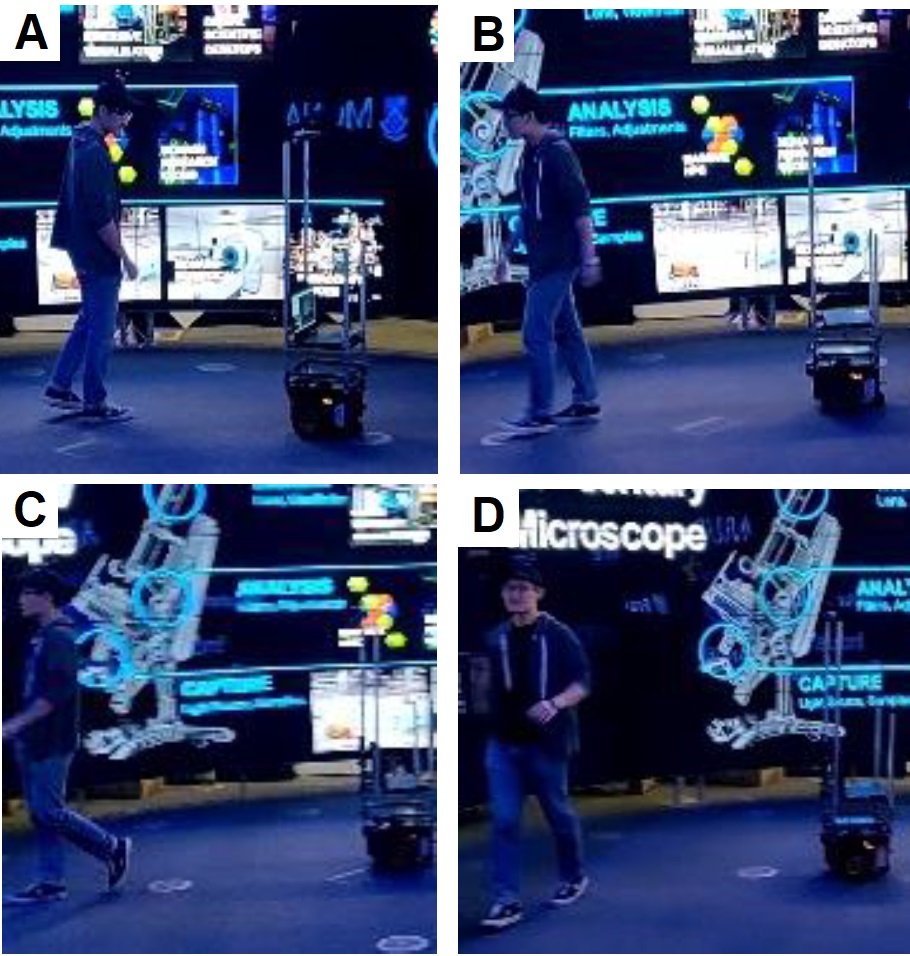}
\caption{Our robot identifying and following a target person in our experiment.}
\label{fig:experiment}
\end{figure}

To compare the performance of the three methods, we measured the average absolute tracking error of all people. We also counted the number of frames where all surrounding people were correctly tracked, the number of frames with incorrectly tracked person(s), and the number of frames with person(s) not detected at all. For Exp 3 and Exp 5, we also counted the number of frames where the target person was correctly track, the number of frames where the target person was incorrectly tracked, and the number of frames where the target person was not detected at all. Furthermore, we computed the CLEAR-MOT metrics, including multiple object tracking precision (MOTP) and multiple object tracking accuracy (MOTA), as proposed by \cite{clearmot}. Figure \ref{fig:experiment} shows a scene of our experiment where the robot is tracking and following a specific target person. We used a simple proportional controller for following the target person in our experiment.

\begin{figure}[t]
\centering
\includegraphics[width=.40\textwidth]{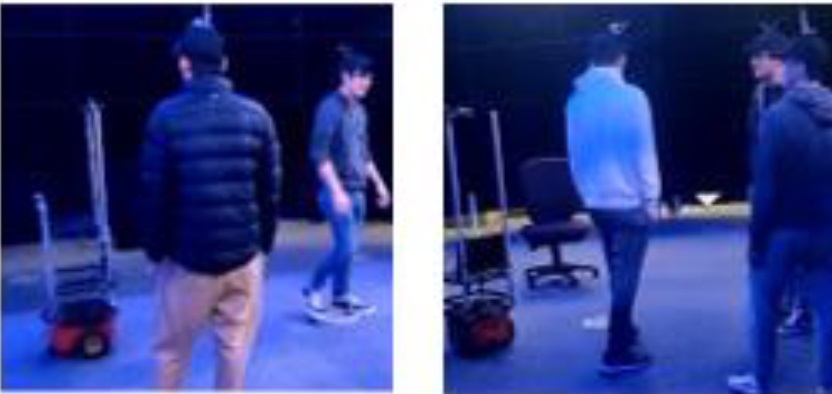}
\caption{Instances in our experiment of people cutting across between the robot and the target person when the robot is following a person.}
\label{fig:experiment-crossing}
\end{figure}

In our experiment scenarios, except the first one, the three people deliberately leave and re-enter the field of view of the robot, occlude each other, and cross paths with the robot on occasions, simulating real-world challenging scenarios. Figure \ref{fig:experiment-crossing} shows instances in the experiment when people cut across in between the robot and the target person it is following. A video illustrating our experiment is available online \footnote{https://youtu.be/FfXNeWVCFxo}.

\section{RESULTS}
        \label{sec:results}
        Table \ref{table:tracking_error} presents the resulting tracking error for each of the three tested methods. Overall, our proposed method yielded the smallest tracking error averaged across all experiment scenarios, demonstrating the superior performance of our proposed algorithm. In fact, in all five experiment scenarios our proposed method yielded error lower than or comparable to the alternative methods.

\begin{figure}[t]
\centering
\includegraphics[width=.48\textwidth]{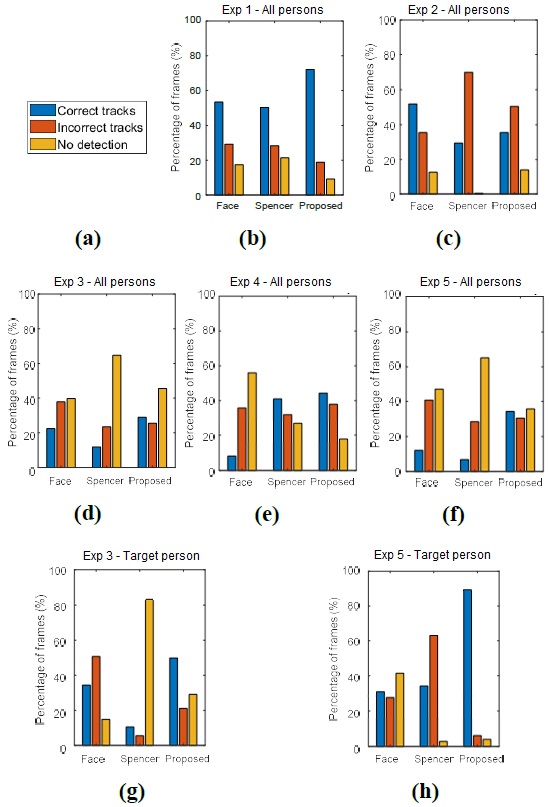}
\caption{Bar charts showing the percentage of frames for each experiment scenario with persons correctly tracked, incorrectly
tracked, and not detected at all. (a): Legend; (b): Exp
1 (c): Exp 2; (d): Exp 3; (e): Exp 4; (f): Exp 5. (g): Exp
3 - target person only, (h): Exp 5 - target person only.}
\label{fig:frame_count_result}
\end{figure}

Figure \ref{fig:frame_count_result} shows the percentage of frames with correctly tracked persons, incorrectly tracked person(s), and un-detected person(s). Note that these results (Figure \ref{fig:frame_count_result}b-f) account for all people surrounding the robot at all times. In most of our experiment scenarios, the persons were deliberately moving in and out of the robot's field of view to create a challenging real-world-like scenario. This results in multiple frames during when there are one or more persons that are not seen by the robot. Hence, one would not expect to observe the percentage of correctly tracked frames to be anywhere near 100\% for these scenarios.

Looking at Figure  \ref{fig:frame_count_result}, the results again show that our proposed algorithm was able to achieve best overall performance. Our proposed algorithm achieved the highest number of frames with correctly tracked persons in all experiment scenarios, except Exp 2. In Exp 2, the robot is stationary while the people are moving around. Hence, this creates the most difficult scenario. Inspecting Figure  \ref{fig:frame_count_result}h showing results for Exp 5, where the robot was following a specific target person, we found that our proposed method was able to achieve far better performance than the alternative methods, correctly tracking the target for approximately 90\% of the time. In fact, despite a few scattered incorrectly tracked frames over the experiment duration, it did not affect the robot's ability in following the target person. Our robot successfully followed the target person until the end of the experiment trial. The alternative methods, on the other hand, failed part way through. This result is particularly promising as Exp 5 represents our intended use case of a robot following a specific user in many human-robot collaborative applications.

The CLEAR-MOT metrics are shown in Table \ref{table:clearmot}. From these results, we see that our proposed method achieved a multiple object tracking precision (MOTP) better than FaceNet, and similar to SPENCER. Furthermore, our proposed method achieved a multiple object tracking accuracy (MOTA) much better than both FaceNet and SPENCER. 

\begin{table}
\caption{Average absolute tracking error for each tracking method in each experiment scenarios.}
\label{table:tracking_error}
\centering
\begin{tabular}{ |c|c|c|c| } 
\hline
\textbf & \textbf{FaceNet} & \textbf{SPENCER} & \begin{tabular}{@{}c@{}}\textbf{Proposed} \\ \textbf{method}\end{tabular} \\
\hline
\textbf{Exp 1} & 1.06m & 0.69m & 0.51m \\ 
\hline
\textbf{Exp 2} & 1.14m & 2.14m & 1.17m \\ 
\hline
\textbf{Exp 3} & 3.87m & 1.86m & 1.52m \\ 
\hline
\textbf{Exp 4} & 9.28m & 2.01m & 1.69m \\ 
\hline
\textbf{Exp 5} & 8.08m & 2.13m & 2.35m \\ 
\hline
\textbf{Average} & 4.68m & 1.77m & 1.47m \\ 
\hline
\end{tabular}
\end{table}

\begin{table}
\caption{Average CLEAR-MOT metrics from all experiment scenarios. }
\label{table:clearmot}
\centering
\begin{tabular}{ |c|c|c|c|c|c| } 
\hline
\textbf & \textbf{FaceNet} & \textbf{SPENCER} & \begin{tabular}{@{}c@{}}\textbf{Proposed} \\ \textbf{method}\end{tabular} \\
\hline
\textbf{MOTP} & 0.288m & 0.185m & 0.216m \\ 
\hline
\textbf{MOTA} & 29.6\% & 23.7\% & 43.0\% \\ 
\hline
\end{tabular}
\end{table}

\section{CONCLUSION}
        \label{sec:conclusion}
        We have presented a method for enabling robots to track and identify surrounding people. Most existing person trackers track people anonymously. Hence, person-following robots developed using these trackers often follow people without knowing who they are following. Our proposed algorithm uses a Sequential Nearest Neighbour with Thresholding Selection algorithm to fuse together data from an anonymous person tracker and a face recogniser to enable identification and tracking of each surrounding person. This allows robots to track and follow specific users, as required in may service robot applications. Experiment results show that our proposed method achieved superior performance compared to alternative methods in terms of tracking error, and percentage of correctly tracked frames. In addition, our experiment scenario of following a specific person demonstrated that our method is robust, and enables a robot to successfully follow a target person continually, when alternative methods fail.

Our method does have a limitation of needing to see the person's face to initially identify a person or re-identify a person in case the person tracker loses track of a person. However, this is similar to how humans identify and track people in many cases. Potential improvements to our method can be incorporating additional features for (re-)identification, such as clothing colour or person's height, or using a more sophisticated model for predicting a person's trajectory when they become occluded or leave the robot's field of view. We will be testing our algorithm in more complex, real-world scenarios in our next steps.

\section{ACKNOWLEDGEMENT}
        \label{sec:acknowledgement}
        This project was supported by the Australian Research Council Discover Projects Grant DP200102858.

\bibliography{main}
\bibliographystyle{IEEEtran}

\end{document}